\documentclass[letterpaper, 10 pt, conference]{ieeeconf}
\IEEEoverridecommandlockouts

\usepackage{booktabs}
\usepackage{cite}
\usepackage{amsmath,amssymb,amsfonts}
\usepackage{float}       
\usepackage{graphicx}
\usepackage{textcomp}
\usepackage{xcolor}
\usepackage{float}
\usepackage{placeins}
\usepackage{booktabs}

\usepackage{hyperref}
\usepackage{multirow}
\usepackage{booktabs}
\usepackage{multirow}
\usepackage{graphicx}
\usepackage{amssymb}
\usepackage{placeins}
\usepackage{dblfloatfix}
\usepackage{capt-of}

\usepackage[acronym]{glossaries}
\newacronym{our_pipeline}{BIT-Nav}{Brain-Inspired Trajectory Memory for Embodied Navigation}
\newacronym{llm}{LLM}{Large Language Model}
\newacronym{gru}{GRU}{Gated Recurrent Unit}
\newacronym{vlm}{VLM}{Vision-Language Model}
\newacronym{vln}{VLN}{Vision-and-Language Navigation}
\newacronym{bite}{BITE}{Bi-GRU Trajectory Encoder}
\makeglossaries

\begin{document}

\title{BIT-Nav: Brain-Inspired Trajectory Memory for Embodied Navigation}

\author{
Rithvik Jonna$^{1}$,
Aakash Gurram$^{2}$,
Man Namgung$^{2}$,
Wyatt Mackey$^{3}$,
and Tinoosh Mohsenin$^{1,2}$\\
$^{1}$Department of Electrical and Computer Engineering \quad
$^{2}$Laboratory for Computational Sensing and Robotics\\
Johns Hopkins University, \quad
$^{3}$DEVCOM Army Research Laboratory
}
\twocolumn[{
  \renewcommand\twocolumn[1][]{#1}
  \maketitle
  \vspace{-10pt}
  \centering
  \includegraphics[width=7.5in]{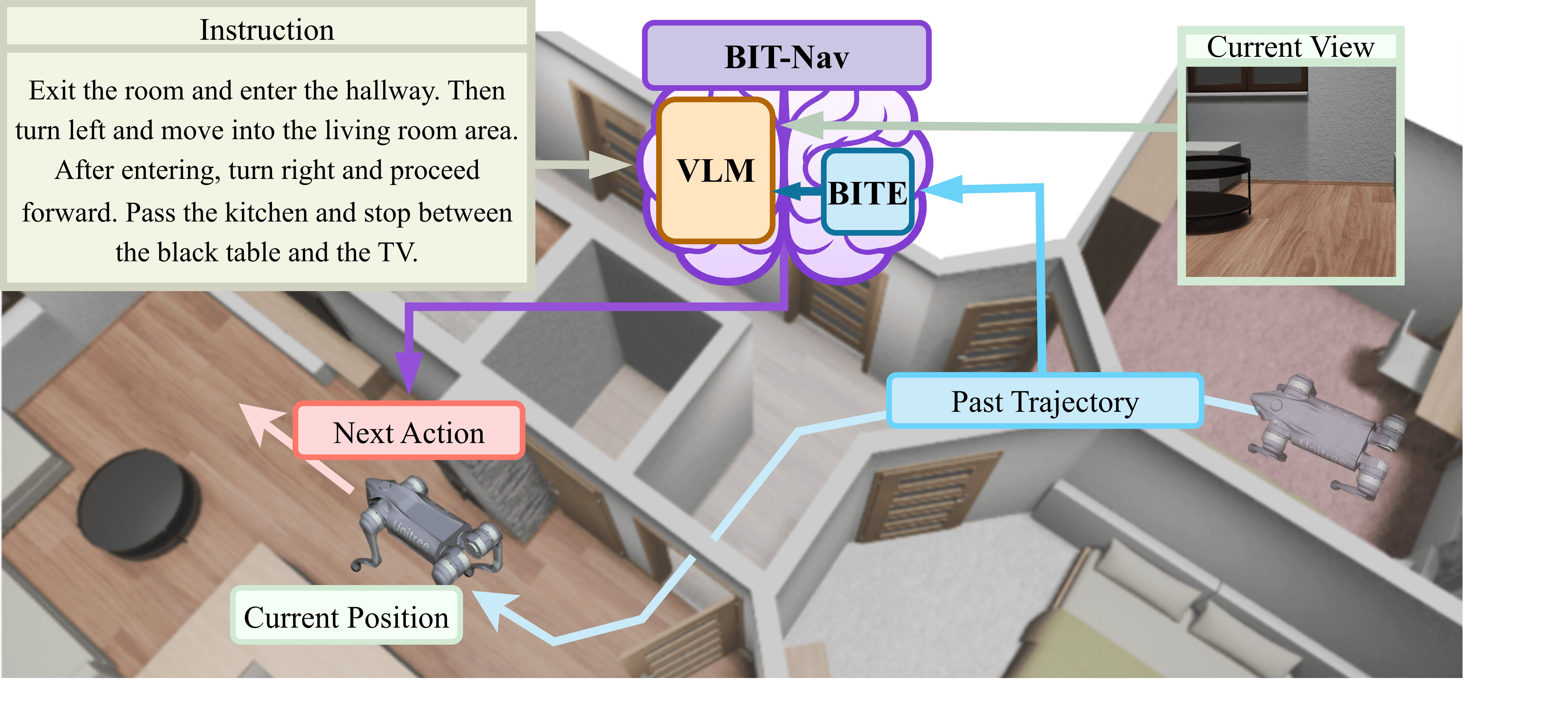}
\vspace{-27pt}
\captionof{figure}{
      \textbf{Brain-Inspired Trajectory Memory for Long-Horizon Embodied Navigation.} \glsentryshort{our_pipeline} consists of two modules: a Vision-Language Model (\glsentryshort{vlm}) and Bi-GRU Trajectory Encoder (\glsentryshort{bite}). The \glsentryshort{vlm} takes the natural-language instruction together with a current camera view, while \glsentryshort{bite} encodes the agent's past motion history into a compact trajectory-memory token and passes it to the \glsentryshort{vlm}. Conditioning jointly on the instruction, the camera views, and the trajectory-memory token, \glsentryshort{our_pipeline} generates the next action.
  }
  \label{fig:01}
  \vspace{1em}
  }]

\begin{abstract}
Vision-Language Action Models (VLAs) have demonstrated promising results 
in embodied navigation, yet existing approaches rely on selecting 
a fixed number of frames from a growing trajectory history. As 
episodes extend, this selection grows increasingly sparse, and 
prior work shows no accuracy gain when scaling from 8 to 64 
frames suggesting that the bottleneck 
is not frame quantity but the nature of the representation itself. 
We hypothesize that sparse frame selection cannot capture the 
structured behavioral signal turning patterns, cumulative 
displacement, and path topology that long-horizon reasoning 
requires.
We introduce \textbf{BIT-Nav} (Brain-Inspired Trajectory Memory 
for Embodied Navigation), a framework that augments existing \gls{vlm} 
navigation pipelines with a compact learned trajectory memory. 
Motivated by hippocampal path integration, where spatial experience 
is compressed into structured episodic traces rather than stored 
as raw sensory replay, 
BIT-Nav trains a GRU-based trajectory encoder over action and 
relative pose sequences via a SimCLR-style contrastive 
objective on augmented trajectory pairs 
sharing the same start and end points. The resulting embedding 
is projected into the VLM token space via a learned linear 
projector and appended as a single memory token at each decision 
step adding negligible overhead regardless of trajectory 
length alongside the current visual observation and sparse 
frame history. We demonstrate that BIT-Nav enables structured
long-horizon reasoning in Vision-and-Language
Navigation settings, achieving 0.83 accuracy on
cumulative heading estimation and matching
history-augmented baselines on action-frequency
inference at 22$\times$ lower token cost where
single frame collapse to ${\approx}0.07$
as trajectories grow beyond 20 steps.

\end{abstract}

\section{Introduction}

The emergence of Vision-Language Models~(VLMs) has 
fundamentally expanded the capabilities of embodied AI 
systems. By unifying visual perception and natural language 
reasoning within a single model, VLMs enable agents to 
interpret complex instructions, ground language in physical 
scenes, and generalize across diverse, previously unseen 
environments~\cite{driess2023palme, brohan2023rt2}. These 
capabilities have made VLMs an increasingly attractive 
backbone for Vision-and-Language Navigation (\gls{vln}), where 
an agent must interpret a natural language command execute a corresponding sequence of actions in an 
unknown environment without access to a 
map~\cite{anderson2018r2r, krantz2020vlnce}. Recent 
VLM-based navigation systems have demonstrated strong 
performance on standard benchmarks, leveraging the broad 
world knowledge and cross-modal reasoning capabilities of 
pretrained foundation models to follow step-by-step 
instructions with increasing 
fidelity~\cite{cheng2024navila, shah2023lm, chen2022hm3d}. 
Yet despite these advances, a fundamental limitation 
remains: these systems are \textit{stateless at the 
reasoning level}, and their ability to handle long-horizon 
tasks where decisions depend critically on what has 
already been done remains substantially under-addressed.

In long-horizon vision-language navigation (VLN), successful navigation
requires more than reacting to the current observation. An agent must
interpret its present sensory input while maintaining useful information
about where it has been, how it has moved, and which parts of the
instruction have already been executed. Prior VLN systems have explored
several mechanisms for incorporating history, including recurrent policy
states, trajectory-level aggregation, and visual-history conditioning.
Recent VLM-based approaches such as NaVILA represent history by sampling
a fixed window of past image frames from the growing trajectory and
providing these frames to the language model as contextual input, captured
at approximately 2 frames per second~\cite{cheng2024navila}.

While this strategy gives the VLM access to past visual observations, it
becomes increasingly limited as episode length grows. A fixed-size visual
history must cover trajectories spanning many meters and many decision
steps, so the temporal sampling becomes sparse and can omit intermediate
motion structure that is important for grounding future decisions.
NaVILA's own ablation results suggest that simply increasing the number
of sampled history frames is not sufficient: scaling the history window
from 8 to 64 frames does not improve navigation accuracy~\cite{cheng2024navila}.
This indicates that the limitation is not only the quantity of historical
frames, but also the form in which trajectory history is represented.

In particular, visual-frame history provides only an indirect account of
the agent's motion. Low-level recurrent states used in control heads or
policy networks may encode aspects of past behavior, but these states are
typically not exposed to the VLM as an explicit, structured, and queryable
trajectory representation. As a result, the language model can reason
richly over the current visual scene and sampled past observations, but
has limited access to a compact representation of the agent's accumulated
motion history. This creates a gap between the perceptual strength of
modern VLMs and their ability to perform temporally grounded reasoning
over long navigation episodes.

This distinction between storing raw appearance frames and encoding
structured motion history has a compelling parallel in biological
navigation. The hippocampal--entorhinal system does not simply archive
sensory observations; it integrates motion, heading, and displacement
signals into compact spatial memories that support retrieval and
planning~\cite{moser2008place}. Recent work further shows that analogous
spatial representations can emerge in neural networks trained for
navigation~\cite{banino2018vector}, suggesting that learned motion
encoding is a computationally principled strategy.
Motivated by this view, we propose \textbf{BIT-Nav}, which treats
trajectory history as an explicit modality for VLM-based navigation.
BIT-Nav trains a GRU-based encoder over robot actions and relative pose
displacements using a SimCLR-style contrastive objective~\cite{chen2020simclr},
so that trajectories with shared start and end states are mapped to
similar embeddings despite local path variation. This embedding is then
projected into the VLM token space and supplied as a memory token
alongside the current visual observation and sampled frame history.
By placing trajectory memory directly inside the VLM reasoning context,
rather than only inside a recurrent control state, BIT-Nav enables the
model to reason over where it has been and which parts of the instruction
remain to be executed. We show that this gives rise to
\textbf{instruction localization}, allowing the model to ground current
decisions in the progress already made through a long-horizon instruction.

\noindent We summarize our contributions as follows:

\begin{itemize}

\item \textbf{Trajectory as an Inherent Navigation Modality.}
We introduce trajectory and position as an essential modality for
vision-language navigation, complementing visual observations and language
instructions with an explicit representation of how the agent moves through
space. We train this trajectory component using contrastive supervision,
encouraging navigation histories with shared intent and endpoints to form
compact, discriminative representations.

\item \textbf{Emergent Instruction Localization.} 
We demonstrate that incorporating trajectory memory 
enables implicit tracking of instruction progress 
without explicit supervision, allowing the model to 
infer completed sub-goals and ground decisions in 
remaining navigation steps.

\item \textbf{Trajectory--VLM Alignment for Long-Horizon 
Reasoning.} We align the learned trajectory embeddings 
with Qwen3-VL-8B~\cite{Qwen3-VL} via a learned 
projector into the VLM token space~\cite{lin2024vila}, 
enabling the model to condition its reasoning on 
structured motion history alongside current visual 
observations within a single forward pass.

\item \textbf{Trajectory Memory Dataset Construction.} 
We construct a trajectory-aware dataset from R2R and 
RxR~\cite{anderson2018r2r, ku2020rxr} by extracting 
structured action and pose sequences from navigation 
episodes. We further introduce an augmentation strategy 
that generates geometrically equivalent but executionally 
diverse trajectory variants, enabling learning of motion 
structure rather than surface-level action sequences.

\end{itemize}
\section{Related Work}

\subsection{Trajectory Representation Learning}

Learning compact representations of agent trajectories 
has been a long-standing problem in robot learning and 
sequential decision-making. Early work on recurrent 
neural networks demonstrated that GRUs and LSTMs can 
encode variable-length action sequences into fixed-size 
hidden states suitable for policy 
conditioning~\cite{hausknecht2015drqn, zhu2017target}. 
More recently, contrastive representation learning has 
been applied to trajectory data to learn embeddings 
that are invariant to observation noise and execution 
variability. CURL~\cite{laskin2020curl} and 
SPR~\cite{schwarzer2020spr} demonstrated that 
self-supervised contrastive objectives over state 
sequences yield representations that transfer 
effectively to downstream control tasks. In the 
context of navigation, trajectory encoders have been 
used to summarize path history for topological 
map construction~\cite{savinov2018semiparametric} 
and for learning waypoint 
predictors~\cite{shah2022viking}. 

Our work differs from these approaches in two key 
respects. First, we train the trajectory encoder 
exclusively on the motion stream discrete action 
tokens and relative pose displacements rather 
than on visual observations, isolating behavioral 
structure from appearance. Second, and most 
critically, we align the learned trajectory embedding 
directly into the token space of a Vision-Language 
Model, making motion history a first-class input to 
language model reasoning rather than a conditioning 
signal for a separate control head.

\subsection{Vision-Language Navigation and 
Vision-Language-Action Models}

\gls{vln} requires an 
agent to follow natural language instructions through 
previously unseen environments, grounding linguistic 
descriptions in visual 
observations~\cite{anderson2018r2r, krantz2020vlnce, 
ku2020rxr}. Early approaches relied on sequence-to-sequence 
models with attention over instruction 
tokens~\cite{anderson2018r2r, fried2018speaker}, 
while subsequent work introduced graph-based 
representations~\cite{chen2022hm3d} and 
transformer-based cross-modal 
grounding~\cite{chen2021hamt, hong2021vln}. 

The emergence of large pretrained Vision-Language 
Models has shifted the paradigm: recent systems 
fine-tune VLMs directly on navigation data, 
leveraging broad world knowledge for 
generalization~\cite{shah2023lm, zhou2024navgpt}. 
NavGPT~\cite{zhou2024navgpt} demonstrates zero-shot 
navigation reasoning using GPT-4 with textual 
observation summaries, while EmbodiedGPT~\cite{mu2024embodiedgpt} 
and PaLM-E~\cite{driess2023palme} extend VLMs to 
embodied planning more broadly. NaVILA~\cite{cheng2024navila} 
proposes a two-level framework in which a VLM 
generates mid-level language actions executed by a 
low-level RL locomotion policy, demonstrating strong 
performance on continuous VLN benchmarks. However, 
the VLM component in NaVILA is invoked fresh at each 
step with a fixed window of uniformly sampled 
historical frames, and prior ablation studies within 
that work show no accuracy improvement when scaling 
the history window from 8 to 64 
frames~\cite{cheng2024navila} revealing that 
appearance-based frame selection is an insufficient 
mechanism for long-horizon memory. 

Vision-Language-Action models~(VLAs) such as 
RT-2~\cite{brohan2023rt2} and 
OpenVLA~\cite{kim2024openvla} extend this paradigm 
to robotic manipulation by fine-tuning VLMs to 
predict low-level actions directly. While these 
models demonstrate impressive generalization in 
manipulation settings, they do not address the 
long-horizon memory problem inherent to extended 
navigation episodes. BIT-Nav is complementary to 
this line of work: rather than replacing the VLM 
action head, we augment the VLM's input context 
with a learned trajectory memory token, preserving 
the VLM's reasoning capability while extending it 
to temporally structured motion history.


\subsection{Memory, Spatial Grounding, and 
Multimodal Alignment}

Equipping neural agents with explicit memory 
mechanisms for long-horizon reasoning has been 
studied extensively. Neural Turing 
Machines~\cite{graves2014ntm} and Differentiable 
Neural Computers~\cite{graves2016dnc} introduced 
external memory with addressable read/write 
operations, while episodic memory architectures 
have been applied to navigation for storing and 
retrieving past 
observations. 
Transformer-based approaches extend context windows 
to handle longer histories, with models such as 
Recurrent Memory Transformers~\cite{bulatov2022rmt} 
and Memory-Augmented Transformers~\cite{wu2022memformer} 
demonstrating improved performance on tasks requiring 
long-range temporal dependencies. In the VLN 
setting, HAMT~\cite{chen2021hamt} encodes full 
trajectory history via a hierarchical attention 
mechanism over past observations, and 
ETPNav~\cite{an2023etpnav} builds topological maps 
from visited nodes to support backtracking and 
re-planning. 

However, these memory mechanisms operate within 
specialized navigation architectures and do not 
transfer motion history into the reasoning context 
of a general-purpose VLM. BIT-Nav addresses this 
gap by encoding trajectory history as a compact 
single token projected directly into the VLM token 
space, making past motion available to the language 
model's attention mechanism without modifying its 
architecture.

Multimodal alignment training a projection 
layer to map representations from one modality 
into the token space of a language model has 
been central to the development of modern VLMs. 
CLIP~\cite{radford2021clip} established contrastive 
vision-language pretraining as a foundation for 
cross-modal alignment, while 
LLaVA~\cite{liu2023llava} and 
VILA~\cite{lin2024vila} demonstrated that a 
lightweight MLP projector trained on 
instruction-following data is sufficient to align 
visual encoders with large language models. 
Flamingo~\cite{alayrac2022flamingo} extended this 
to interleaved vision-language sequences via 
cross-attention, and InstructBLIP~\cite{dai2023instructblip} 
introduced query-based alignment via a Q-Former 
module. BIT-Nav follows this alignment paradigm 
but introduces a novel modality: rather than 
aligning a visual encoder, we align a learned 
trajectory encoder trained on motion sequences 
rather than images into the token space of 
Qwen3-VL-8B~\cite{Qwen3-VL}, establishing 
trajectory memory as a queryable modality within 
the VLM's reasoning context.

\section{Methodology}
\label{sec:methodology}

We consider the problem of learning a compact trajectory representation from discrete navigation action sequences, such that the resulting embedding can serve as a trajectory memory token consumed by a downstream large language model. Our pipeline consists of four stages: (i) converting raw discrete actions into structured 7-D geometric motion features, (ii) arranging these sequences into group-aware contrastive minibatches, (iii) encoding each variable-length sequence with a bidirectional GRU Trajectory Encoder [\textbf{BITE block}] trained under a multi-positive InfoNCE objective, and (iv) passing the frozen encoder's output through a learned lightweight multimodal projector so that the trajectory memory becomes available to the LLM (Qwen3-VL-8B) at inference time. During training, positive and negative batches teach the Bi-GRU to behave as a trajectory memory unit. During inference, real-time trajectories are encoded on the fly and the resulting embedding is injected into the LLM, giving it structured access to the agent's navigation history.

\begin{figure*}[!t]
  \centering
  \includegraphics[width=\textwidth]{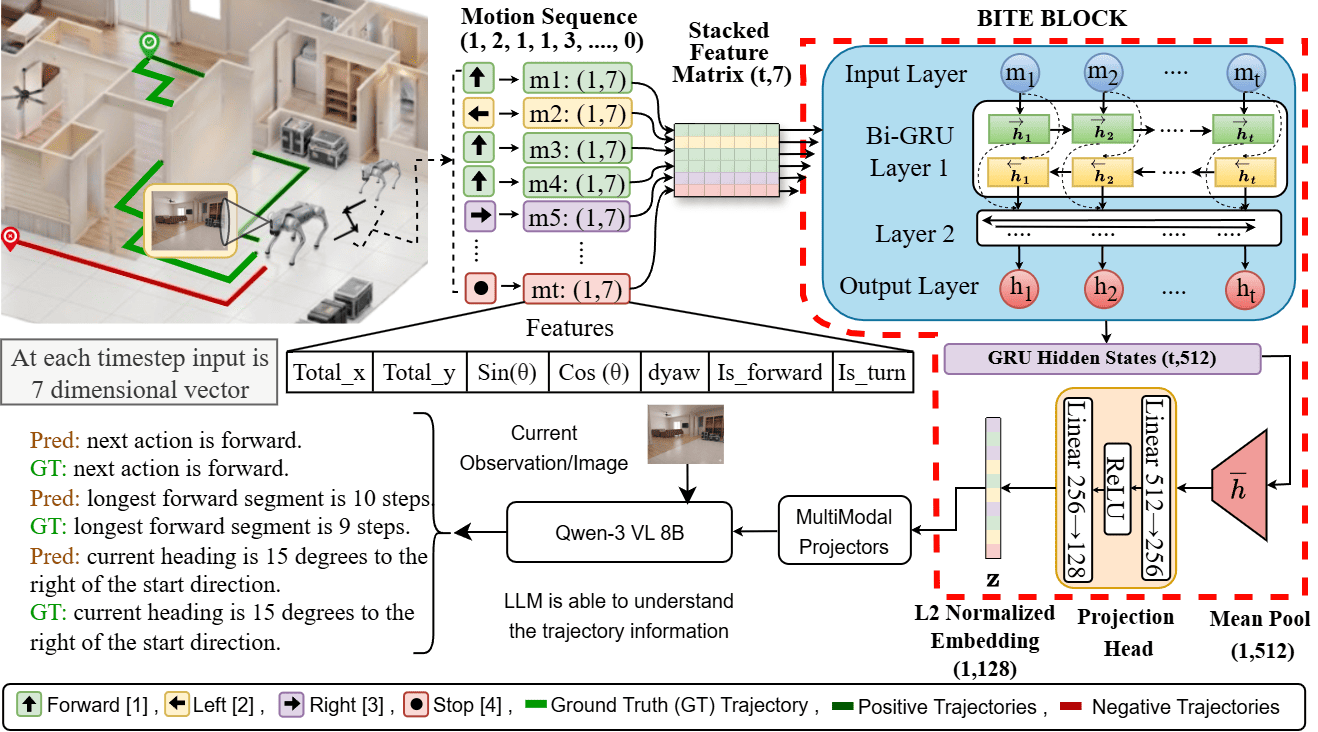}
  \caption{\textbf{BITE block.}
The BITE block consists of a 2-layer bidirectional 
GRU that processes the agent's action history as 
7-D motion features $m_{1:t}$, a masked mean-pool 
that compresses the full sequence into a single 
vector $\bar{h} \in \mathbb{R}^{512}$, and a 
multimodal projector MLP that maps $\bar{h}$ into 
the token space of Qwen3-VL-8B - adding exactly 
one trajectory memory token to the LLM input at 
every decision step, regardless of trajectory 
length.}
  \label{fig:overiew}
\end{figure*}

\subsection{Trajectory Representation and Feature Engineering}
\label{subsec:features}

Let a navigation trajectory be a sequence of discrete actions $a_{1:T}$, where each action is drawn from the alphabet
\begin{equation}
\mathcal{A} = \{\text{stop},\ \text{forward},\ \text{turn-left},\ \text{turn-right}\}.
\label{eq:action_alphabet}
\end{equation}

Rather than feeding raw action IDs into the encoder, we represent each action into a 7-dimensional motion feature vector that exposes both the local control signal and the accumulated geometric state of the agent. This enriched representation makes position, heading, and motion type explicitly available to the recurrent encoder.

At each timestep $t$, we define
\begin{equation}
m_t = \bigl[\, c^{x}_t,\ c^{y}_t,\ \sin\theta_t,\ \cos\theta_t,\ \Delta\theta_t,\ f^{\text{fwd}}_t,\ f^{\text{turn}}_t \,\bigr] \in \mathbb{R}^{7},
\label{eq:motion_feature}
\end{equation}
where $(c^{x}_t, c^{y}_t)$ is the cumulative planar position in the local frame, $\theta_t$ is the accumulated heading angle, $\Delta\theta_t$ is the heading change at the current step, and $f^{\text{fwd}}_t, f^{\text{turn}}_t \in \{0,1\}$ are indicator variables identifying whether the current action is a forward translation or a rotation.

The geometry is updated using a fixed step length $d = 0.25\,\text{m}$ for forward motion and a fixed turn angle $\alpha = 15^{\circ}$ for rotations. The heading increment is defined as
\begin{equation}
\Delta\theta_t =
\begin{cases}
+\alpha, & a_t = \text{turn-left}, \\
-\alpha, & a_t = \text{turn-right}, \\
0, & \text{otherwise},
\end{cases}
\label{eq:dtheta}
\end{equation}
and the accumulated heading and position evolve as
\begin{align}
\theta_t &= \theta_{t-1} + \Delta\theta_t, \label{eq:theta_update} \\
c^{x}_t &= c^{x}_{t-1} + d\cdot\mathbb{I}(a_t = \text{forward})\cos\theta_t, \label{eq:cx_update} \\
c^{y}_t &= c^{y}_{t-1} + d\cdot\mathbb{I}(a_t = \text{forward})\sin\theta_t. \label{eq:cy_update}
\end{align}

The full trajectory is then represented as the variable-length feature sequence $m_{1:T} = (m_1, \dots, m_T)$ with $m_t \in \mathbb{R}^{7}$. This construction preserves the original action semantics through the forward and turn indicators, while simultaneously exposing the encoder to cumulative position and heading, which distinguishes action fragments that look similar locally but correspond to very different global motion patterns.

\subsection{Batch Construction for Contrastive Training}
\label{subsec:batching}

The dataset is organized at the level of start-goal pairs. It contains $M = 3600$ groups, each corresponding to a distinct start-goal pair, and each group holds exactly 10 augmented trajectories that connect the same source and destination, yielding $36{,}000$ trajectories in total. Trajectories within a group share endpoints but may differ in step composition, intermediate turns, and total length, so the encoder is pushed to learn the broader structural signature of a start-goal episode rather than memorize a single canonical path.

To exploit this group structure during training, we adopt a group-aware batch sampler. Each minibatch is constructed by sampling $B_G$ complete groups and including all 10 trajectories from each sampled group, giving a batch of $B = 10\,B_G$ trajectory-feature sequences. In our setup we use $B_G = 6$ groups per batch, which yields
\begin{equation}
B = 10 \times 6 = 60 \quad \text{trajectories per minibatch}.
\label{eq:batch_size}
\end{equation}

Because every group always contributes all 10 of its trajectories, each anchor in the batch is guaranteed to see
\begin{equation}
\begin{aligned}
|\mathcal{P}(i)| &= 9, \quad \text{positives},\\
|\mathcal{N}(i)| &= 10(B_G - 1) = 50, \quad \text{negatives},
\end{aligned}
\label{eq:pos_neg_counts}
\end{equation}
without any need for hard-negative mining or an external memory bank. Variable-length sequences within a batch are padded to the length of the longest sequence, a length mask is retained for later pooling, and the resulting tensor is passed to the Bi-GRU encoder.

\subsection{BITE (Bi-GRU Trajectory Encoder) Block}
\label{subsec:encoder}

The encoder $E_{\tau}$ maps a variable-length motion sequence $m_{1:T}$ to a fixed-dimensional embedding $z \in \mathbb{R}^{d}$. It is built from a 2-layer bidirectional GRU followed by a small projection head.

At each timestep, the bidirectional GRU produces a hidden state by concatenating a forward and a backward state:
\begin{equation}
h_t = \bigl[\,\overrightarrow{h}_t\,;\,\overleftarrow{h}_t\,\bigr] \in \mathbb{R}^{2 d_h},
\label{eq:bigru_hidden}
\end{equation}
with hidden dimension $d_h = 256$ per direction, giving a per-step output of size $512$. Rather than using only the final hidden state, we aggregate the full recurrent sequence with a masked mean-pool over the valid (non-padded) timesteps:
\begin{equation}
\bar{h} = \frac{1}{T}\sum_{t=1}^{T} h_t \in \mathbb{R}^{512}.
\label{eq:mean_pool}
\end{equation}

Mean pooling makes the embedding a summary of the entire path rather than a snapshot of the final state, which is particularly important for longer trajectories where meaningful structure appears throughout the sequence.

The pooled representation is then passed through a projection head with a single hidden layer and ReLU activation,
\begin{equation}
u = W_2\,\sigma(W_1\,\bar{h} + b_1) + b_2,
\label{eq:projection}
\end{equation}
mapping $512 \rightarrow 256 \rightarrow 128$, and the final embedding is obtained by $\ell_2$ normalization,
\begin{equation}
z = \frac{u}{\lVert u \rVert_2} \in \mathbb{R}^{128}.
\label{eq:l2_norm}
\end{equation}

Normalization places all trajectory embeddings on the unit hypersphere, so cosine similarity becomes the natural scoring function for both the training objective and downstream retrieval.

\subsection{Multi-Positive InfoNCE Objective}
\label{subsec:infonce}

The encoder is trained with a multi-positive InfoNCE objective computed over the group-structured minibatch. For two embeddings $z_i$ and $z_j$ in the batch, we define the scaled cosine similarity
\begin{equation}
S_{ij} = \frac{z_i^{\top} z_j}{\tau},
\label{eq:similarity}
\end{equation}
with temperature $\tau = 0.07$. Because the embeddings are $\ell_2$-normalized, the inner product is equivalent to cosine similarity. The diagonal of the similarity matrix is masked out ($S_{ii} = -\infty$) to prevent trivial self-matching.

For each anchor $i$ in the batch, the positive set $\mathcal{P}(i) = \{j : y_j = y_i,\ j \neq i\}$ contains all 9 other trajectories from the same start-goal group, while the negative set $\mathcal{N}(i) = \{k : y_k \neq y_i\}$ contains the remaining 50 trajectories in the batch. The per-anchor loss averages the log-probability of each positive over the full non-self denominator:
\begin{equation}
\mathcal{L}_i = -\,\frac{1}{|\mathcal{P}(i)|}\sum_{j \in \mathcal{P}(i)} \log
\frac{\exp\!\bigl(z_i^{\top} z_j / \tau\bigr)}
     {\sum_{k \neq i}\exp\!\bigl(z_i^{\top} z_k / \tau\bigr)},
\label{eq:per_anchor_loss}
\end{equation}
and the batch loss is the mean over all anchors:
\begin{equation}
\mathcal{L}_{\text{InfoNCE}} = \frac{1}{B}\sum_{i=1}^{B}\mathcal{L}_i.
\label{eq:batch_loss}
\end{equation}

The multi-positive formulation is central to our design. A conventional single-positive contrastive loss would only pair an anchor with one other trajectory. By instead averaging over all 9 intra-group positives, the encoder is pushed to compress the entire start-goal equivalence class into a compact neighborhood on the hypersphere. This is precisely what turns the Bi-GRU into a trajectory memory unit: the learned embedding captures what is structurally invariant across all trajectories that solve the same navigation instance, rather than latching onto any single example.

\subsection{Integration with Qwen3-VL-8B}
\label{subsec:llm_integration}

Streaming trajectory encoding.
Once trained under the multi-positive InfoNCE objective,
the Bi-GRU encoder is frozen and operates in a streaming
fashion at inference. At each step~$t$, the accumulated
action history is encoded into the 7-D motion-feature
sequence $m_{1:t}$ and mean-pooled (Eq.~\eqref{eq:mean_pool})
to produce a trajectory memory vector
$z_t = \bar{h}_t \in \mathbb{R}^{128}$,
summarizing the complete action history from episode
start to step~$t$.

Multimodal projector.
A lightweight two-layer MLP with GELU
activation~\cite{hendrycks2016gaussian} maps $z_t$
into the 4096-D token space of Qwen3-VL-8B:
\begin{equation}
  e_{\text{traj}}
    = W_2\,\sigma\!\bigl(W_1\,z_t + b_1\bigr) + b_2,
  \label{eq:projector}
\end{equation}
where $W_1\!\in\!\mathbb{R}^{1024\times256}$ and
$W_2\!\in\!\mathbb{R}^{4096\times1024}$.
This projector introduces only 4{,}461{,}568 parameters
($<$0.06\% of total model size) and is the \emph{only}
component trained during alignment.

Frozen components.
Both the Bi-GRU encoder (302{,}208 parameters) and
Qwen3-VL-8B ($8.77{\times}10^{9}$ parameters) are
kept fully frozen throughout alignment training.
Freezing the encoder preserves contrastively learned
trajectory structure; freezing the LLM preserves its
vision-language reasoning capability. The projector
acts as a learned bridge between the two frozen
representations.

Token injection and context layout.
The projected vector $e_{\text{traj}}\!\in\!\mathbb{R}^{4096}$
is inserted as a single soft token prepended to the
LLM input at each decision step:
\begin{equation}
  \mathbf{X}_t =
  \bigl[
    e_{\text{traj}},\;
    e_{\text{visual}},\;
    e_{\text{frames}},\;
    e_{\text{instr}}
  \bigr],
  \label{eq:input_sequence}
\end{equation}
where $e_{\text{visual}}$ is the current observation
embedding, $e_{\text{frames}}$ are sparse historical
frame tokens, and $e_{\text{instr}}$ is the tokenized
instruction. Prepending $e_{\text{traj}}$ ensures all
self-attention layers attend jointly to structured
motion history, visual perception, and language.

Efficiency.
This design costs exactly \emph{one token} regardless
of trajectory length. As episodes extend and the frame
history window grows increasingly sparse,
$e_{\text{traj}}$ continues to carry dense, structured
context derived from the \emph{complete} action history.
This fixed-size representation supports
trajectory-grounded reasoning, including next-action
prediction, heading estimation, motion segmentation,
and sub-instruction completion inference
(Fig.~\ref{fig:exp_setup}).


\section{Results}
\label{sec:results}

\subsection{Ablation Study and Design Decisions}
\label{subsec:ablation}

To justify the design of the trajectory encoder used in BIT-Nav, we conduct a series of ablation studies analyzing architectural choices, input representations, and contrastive learning objectives. These experiments aim to identify a configuration that not only performs well in trajectory retrieval but also produces embeddings suitable for downstream alignment with a \gls{vlm}.

\subsubsection{Architecture Selection}

\begin{figure}[!htbp]
  \centering
  \includegraphics[width=\columnwidth]{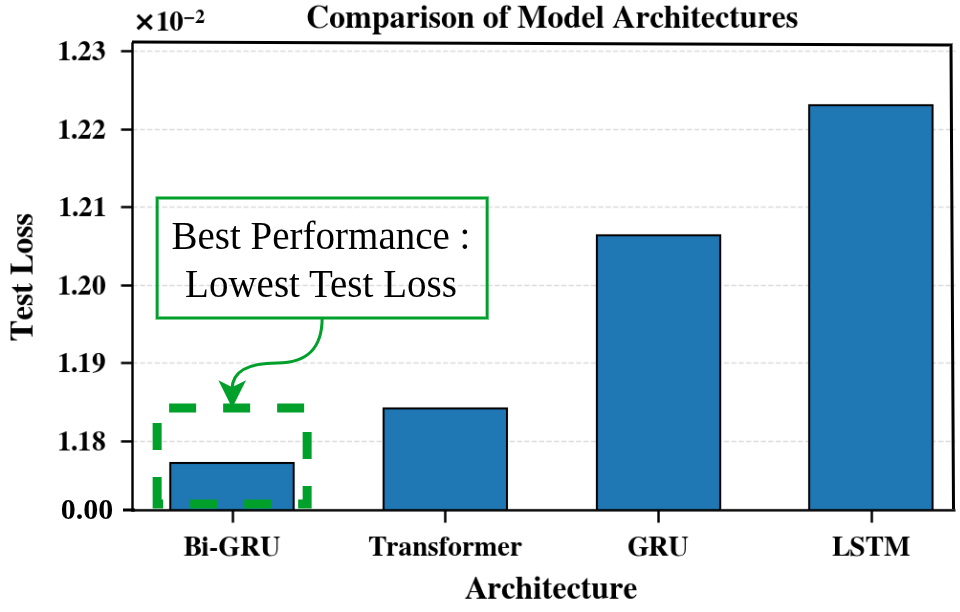}
  \caption{Architecture comparison for trajectory encoding. Bi-GRU achieves the lowest test loss compared to Transformer, GRU, and LSTM.}
  \label{fig:arch_comp}
\end{figure}

We compare multiple sequence modeling architectures for trajectory encoding, including Bi-GRU, Transformer, GRU, and LSTM. As shown in Fig.~\ref{fig:arch_comp}, Bi-GRU achieves the lowest test loss among all evaluated models.

This choice is also consistent with prior sequence-modeling work. Zhao
et al.~\cite{zhao2025cnn_bigru} adopt a CNN-Bi-GRU model for time-series
forecasting and compare it against ARIMA, standalone GRU, and EEMD-ARIMA
baselines, demonstrating improved forecasting performance. Motivated by this
evidence and our own empirical comparison, we adopt Bi-GRU as the core
trajectory encoder in \gls{bite}. Its bidirectional structure allows the encoder
to capture complete sequence-level context, making it well suited for learning
compact trajectory representations from navigation histories.

\FloatBarrier

\subsection{SimCLR-Based Representation Learning}

\begin{figure}[!htbp]
    \centering
    \includegraphics[width=\columnwidth]{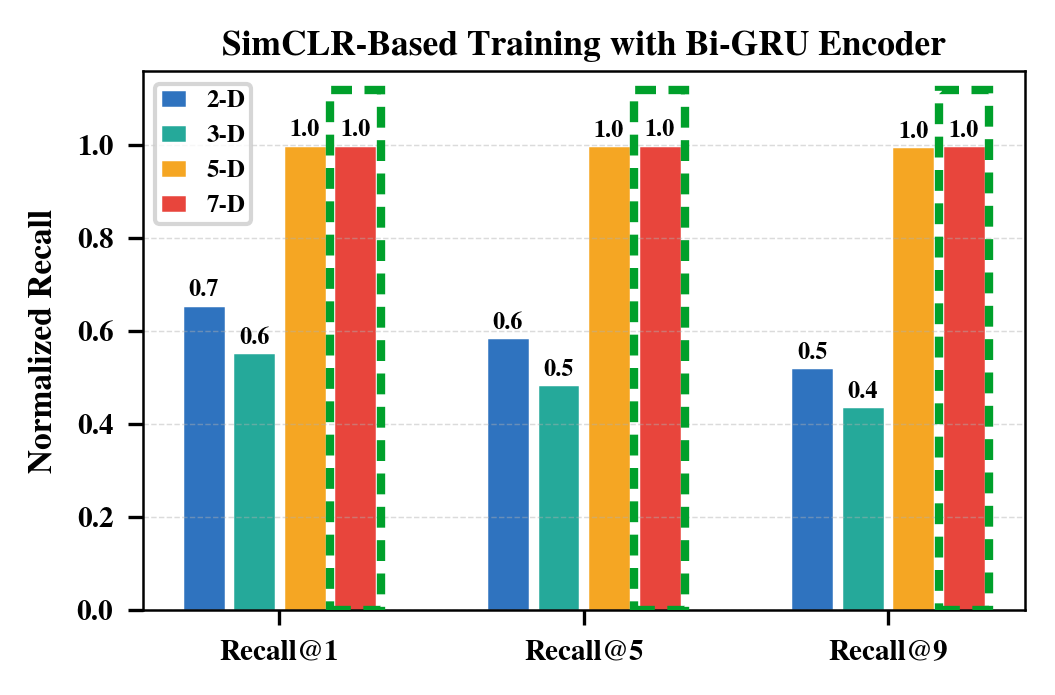}
    \vspace{-1.7em}
    \caption{SimCLR-based trajectory retrieval using a Bi-GRU encoder with different input feature dimensions. The 2-D input consists of $[\texttt{is\_forward},\texttt{is\_turn}]$; the 3-D input uses $[\Delta x,\sin\theta,\cos\theta]$; the 5-D input uses $[c_x,c_y,\Delta\theta,\texttt{is\_forward},\texttt{is\_turn}]$; and the 7-D input uses $[c_x,c_y,\sin\theta,\cos\theta,\Delta\theta,\texttt{is\_forward},\texttt{is\_turn}]$.}
    \label{fig:simclr_retrieval}
    \vspace{-1.0em}
\end{figure}

We evaluate SimCLR-style contrastive learning with an InfoNCE loss using Bi-GRU encoder across multiple input representations. As shown in Fig.~\ref{fig:simclr_retrieval}, we compare 2-D, 3-D, 5-D, and 7-D motion features using Recall@K.

We assess the quality of the learned trajectory embeddings using nearest-neighbor retrieval under cosine similarity. For a trajectory $i$, let $\mathrm{TopK}(i)$ denote the set of $K$ most similar trajectories in the retrieval pool. Since each trajectory has exactly 9 positives, as defined in Eq.~(8), Recall@K is computed as
\begin{equation}
\mathrm{Recall@K}(i)
=
\frac{1}{9}
\sum_{j \in \mathrm{TopK}(i)}
\mathbb{I}(y_j = y_i),
\end{equation}
and averaged over all query trajectories as
\begin{equation}
\mathrm{Recall@K}
=
\frac{1}{N}
\sum_{i=1}^{N}
\mathrm{Recall@K}(i).
\end{equation}

We report Recall@1, Recall@5, and Recall@9. Recall@1 measures nearest-neighbor precision, while Recall@9 evaluates recovery of the full positive set, since each query has exactly nine positives. Recall@5 is reported only as an intermediate operating point between nearest-neighbor retrieval and full positive-set recovery.

From Fig.~5, we observe that representations based only on limited motion information perform poorly. The 2-D features, which encode only action types, fail to capture meaningful trajectory structure. The 3-D features introduce local motion signals, but still lack sufficient global context for reliable discrimination. In contrast, the 5-D representation, derived from the cumulative trajectory formulation in Section~III, significantly improves retrieval performance by incorporating global position and motion evolution.

The 7-D representation achieves nearly identical retrieval performance to the 5-D while providing a more complete geometric description. Specifically, augmenting the 5-D features with $\sin\theta$ and $\cos\theta$ enables the model to explicitly encode heading orientation, rather than relying solely on incremental heading changes $\Delta\theta$. This allows the encoder 


\begin{table*}[t!]
\caption{Accuracy across trajectory lengths}
\resizebox{\textwidth}{!}{%
\begin{tabular}{lccccccccccccccc}
\hline
Steps                       & \multicolumn{3}{c}{15-35}                & \multicolumn{3}{c}{35-55}                & \multicolumn{3}{c}{55-75}                & \multicolumn{3}{c}{75-100}               & \multicolumn{3}{c}{100-125}              \\ \hline
                            & BITE          & w/o Hist. & w/ Hist. & BITE          & w/o Hist. & w/ Hist. & BITE          & w/o Hist. & w/ Hist. & BITE          & w/o Hist. & w/ Hist. & BITE          & w/o Hist. & w/ Hist. \\
Q1                          & 0.76          & 0.07        & 0.09       & 0.85          & 0.07        & 0.07       & 0.85          & 0.06        & 0.06       & 0.79          & 0.09        & 0.09       & 0.87          & 0.05        & 0.05       \\
Q2                          & 0.53          & 0.34        & 0.68       & 0.45          & 0.25        & 0.43       & 0.44          & 0.34        & 0.42       & 0.42          & 0.45        & 0.48       & 0.45          & 0.36        & 0.37       \\ \hline
Mean                        & \textbf{0.64} & 0.21        & 0.38       & \textbf{0.65} & 0.16        & 0.25       & \textbf{0.65} & 0.2         & 0.24       & \textbf{0.61} & 0.27        & 0.29       & \textbf{0.66} & 0.21        & 0.21       \\ \hline
\end{tabular}%
}
\label{tab:viewpoint_quality}
\end{table*}
\FloatBarrier

to reason about both relative and absolute orientation, improving robustness in cases where trajectories share similar displacement but differ in heading evolution .

Based on these observations, we adopt SimCLR-style contrastive learning with 7-D motion features for the BITE block, as it provides the most expressive and geometrically grounded trajectory representation.
This configuration provides the best trade-off between retrieval performance and alignment capability with the Vision-Language Model.


\subsection{Trajectory--Language Alignment Evaluation}
\label{subsec:alignment_eval}

We evaluate the aligned model on two complementary
axes: (1) quantitative accuracy on trajectory-grounded
queries across varying episode lengths, and (2)
qualitative reasoning quality on long-horizon
navigation episodes in Isaac Sim.

\begin{figure}[!htbp]
    \centering
    \includegraphics[width=\columnwidth]{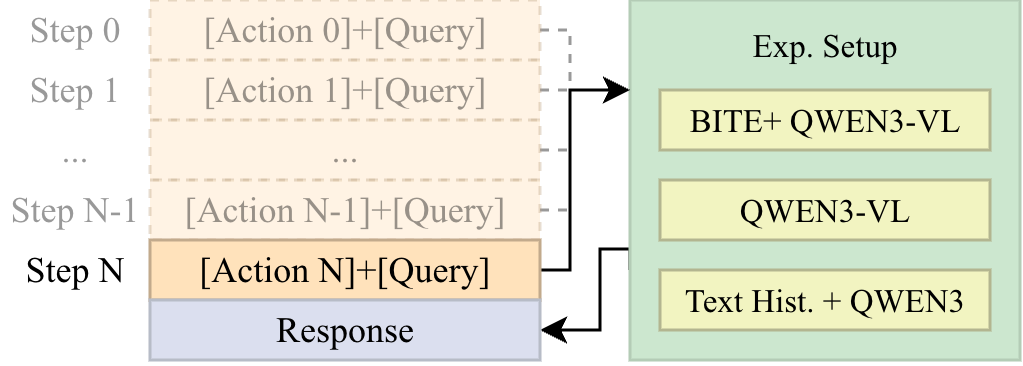}
    \caption{\textbf{Experiment Setup.} We compare three different setups: BITE+Qwen3-VL, vanilla Qwen3-VL, and vanilla Qwen3-VL with text history. At each step, BITE encodes the previous action and the text-history baseline concatenates all past action strings directly into the context.}
    \label{fig:exp_setup}

\end{figure}

\noindent\textbf{Setup:}
We test on three different R2R Episodes using \textbf{BITE+Qwen3-VL},
which encodes the full action history into a single
\texttt{<gru>} token; \textbf{Qwen3-VL}, which
receives no history; and \textbf{Text\,Hist.+Qwen3-VL},
which concatenates all past action strings into the
context. At each step, the model receives the current
\texttt{[Action\,N]+[Query]} pair and generates a
response.
\vspace{0.25em}

\noindent\textbf{Quantitative evaluation.}

We evaluate BITE in token efficiency and accuracy with two queries on randomly generated trajectories whose lengths range from 20 to 124 in steps of 1, with 10 episodes for each length. This range covers trajectories within 3$\sigma$ of the mean length in the R2R dataset. Fig.~\ref{fig:exp_setup} shows the experiment setup and workflow for the evaluation. At every step in each trajectory, BITE encodes the previous action and one context window maintains a full action 
history for vanilla Qwen. With all steps in the trajectory encoded, we compare the model's answer to each query with the ground truth.

Table~\ref{tab:viewpoint_quality} reports binned accuracy across five step ranges. Qwen with BITE achieves a mean of 0.61--0.66 across all ranges, outperforming Qwen with text history (0.21--0.38) and
Qwen without history (0.16--0.27). The Q1 gap is most pronounced: BITE scores 0.76--0.87 while both baselines remain below 0.10 throughout. Overall, Qwen with BITE achieves 0.65 accuracy across all samples, compared to 0.26 for Qwen with text history and 0.21 for Qwen without history.

\begin{figure}[!htbp]
  \centering
  \includegraphics[width=\columnwidth]{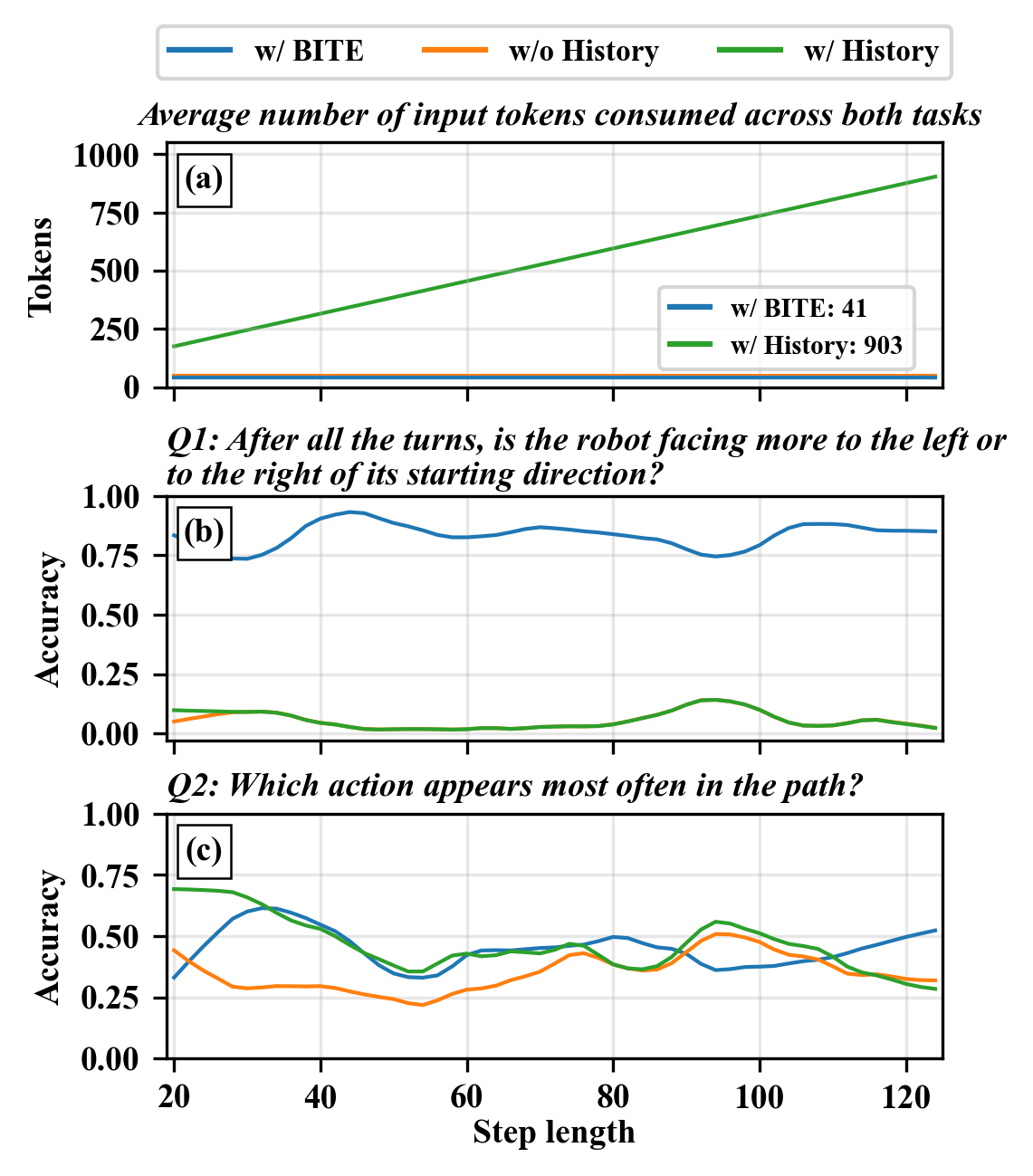}
    \vspace{-2em}
    \caption{\textbf{Evaluation across trajectory lengths.} (a) Average input token count per query across both tasks. Legend values are token counts at step 124. (b–c) Accuracy on Q1 and Q2, respectively. Curves are LOWESS-smoothed (bandwidth 0.20)}
  \label{fig:abla_plot}
\end{figure}

\noindent\textit{Token efficiency.}

As shown in Fig.~\ref{fig:abla_plot}a, the token count of Qwen with text history grows linearly with trajectory length. Instead, BITE injects a single token regardless of history length. With 124 steps, Qwen with text history uses 903 tokens while Qwen with BITE uses only 41 tokens, just 4.5\% of the token used by the Qwen with text history, yet retains dense motion context from the complete action history.

\noindent\textit{Q1: After all the turns,
  is the robot facing more to the left or to the right of its starting direction?}
  
Fig.~\ref{fig:abla_plot}b) shows that Qwen with BITE maintains 0.83 accuracy across all
trajectory lengths.
 \begin{figure*}[t]
    \centering
    \includegraphics[width=\textwidth]{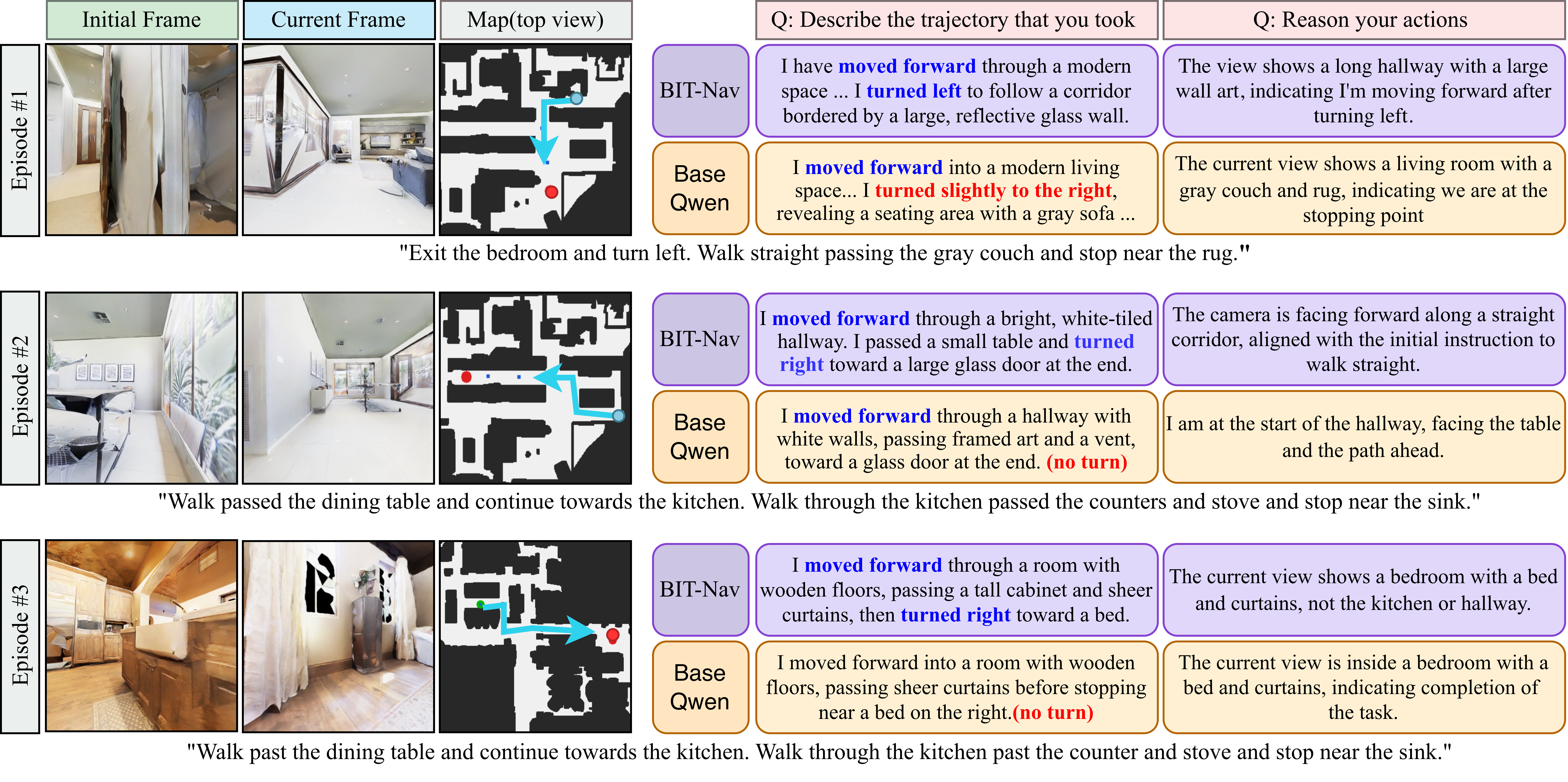}
    \vspace{-15pt}
    \caption{\textbf{Simulation examples of the R2R dataset in Isaac Sim.} The first column shows the initial frame, and the second shows the current frame. The third column indicates the agent’s position on the full map; the blue line represents the agent’s trajectory, and the end of the blue arrow indicates the agent’s current position. Columns 5 and 6 present the model’s responses at the current position. Compared to the base Qwen model, BIT-Nav better captures past trajectory information.}
    \label{fig:01}
\end{figure*}

\FloatBarrier

Both baselines collapse to
${\approx}0.07$, since the Qwen keeps generating ''same'' regardless of input. This confirms that tracking cumulative heading direction requires integrating signed turn angles over the full history, which is lost without BITE.

\noindent\textit{Q2: Which action appears most often in the path?}

According to Fig.~\ref{fig:abla_plot}c, Qwen with BITE and Qwen with text history perform comparably, both outperforming Qwen without history. However, Qwen with BITE matches the text history baseline while keeping token count constant regardless of trajectory length.

\noindent\textbf{Qualitative evaluation
(Fig.~\ref{fig:01}).}
We probe BIT-Nav and Base Qwen3-VL-8B on trajectory
description and action reasoning across three
long-horizon R2R--Isaac episodes.
On trajectory description, BIT-Nav correctly reports
executed turns direction and sequence in all three
episodes. Base Qwen omits turns in Episodes~2 and~3
({\textbf{no turn}}): its descriptions
are visually coherent but historically incorrect,
grounded in the current frame rather than the full
action history.

On action reasoning, the failure persists. Base Qwen
anchors to static scene features and misjudges episode
progress; in Episode~1 it infers a stop near a sofa
despite the agent having turned left and continued
forward. BIT-Nav correctly identifies heading change
and sub-instruction progress.

Both failure modes share a common cause: without
trajectory memory, the model reconstructs history
from the current observation alone, producing outputs
that are perceptually plausible but factually
inconsistent with the executed path.

\subsection{Discussion and Limitations}
\label{subsec:discussion}

The results above establish three properties of BIT-Nav: (i) BITE learns compact, retrievable representations of navigation episodes from motion signals alone; (ii) these representations align successfully to a frozen VLM via a lightweight projector, enabling structured trajectory reasoning; and (iii) this alignment gives rise to instruction localization as an emergent capability without dedicated tracking supervision.

We note that quantitative evaluation on standard VLN benchmarks specifically success rate, SPL, and navigation error on R2R and RxR val-unseen splits against NaVILA baselines with varying frame history sizes~\cite{cheng2024navila} remains as ongoing work. The current results characterize the learned representations and their alignment properties. Full end-to-end benchmark evaluation constitutes the primary next step toward complete empirical validation of BIT-Nav.

\section{Conclusion and Future Work}

We presented \gls{our_pipeline}, a trajectory-aware memory framework for long-horizon vision-and-language navigation. We augment a strong RGB-only VLN model with explicit motion-conditioned memory so that the agent can better preserve state over extended instruction-following episodes. Across standard unseen benchmarks episodes, generalization settings, and ablations, our results are designed to test whether this structured memory improves task tacking and instruction localization under realistic partial observability and long-horizon reasoning demands.

A key direction for future work is to extend this memory beyond trajectory encoding alone and unify it with richer multimodal state representations, including visual landmarks, language subgoals, and uncertainty-aware action selection. Another promising direction is real-robot deployment, where persistent memory may be even more important due to sensor noise, actuation drift, and distribution shift.

\section{Acknowledgment}
This project was sponsored by the U.S. Army Research Laboratory under Cooperative Agreement Number W911NF2120211.
\bibliography{refs/references,refs/eehpc}
\bibliographystyle{ieeetr}
\end{document}